\documentclass[11pt,a4paper]{article}
\usepackage[hyperref]{emnlp-ijcnlp-2019}
\usepackage{times}
\usepackage{latexsym}

\usepackage{url}

\newcommand{\keypoint}[1]{\vspace{0.1cm}\noindent\textbf{#1}\quad}
\usepackage{amssymb}
\usepackage{amsmath}
\usepackage{tikz}
\usepackage{xcolor}
\usepackage[utf8]{inputenc}
\usepackage[english]{babel}
\usepackage{amsthm}
\newtheorem{theorem}{Theorem}

\usepackage{caption}
\usepackage{subcaption}
\usepackage{booktabs}
\usepackage{graphicx}
\usepackage{float}

\aclfinalcopy 

\title{TuckER: Tensor Factorization for Knowledge Graph Completion}

\author{\quad Ivana Bala\v{z}evi\'c$^{1}$ \hspace{1.2cm} Carl Allen$^{1}$ \hspace{.7cm} Timothy M. Hospedales$^{1,2}$\\
  $^1$ School of Informatics, University of Edinburgh, UK\\
  $^2$ Samsung AI Centre, Cambridge, UK\\
  {\tt \{ivana.balazevic, carl.allen, t.hospedales\}@ed.ac.uk} 
  }

\date{}

\begin{document}
\maketitle
\begin{abstract}
  Knowledge graphs are structured representations of real world facts. However, they typically contain only a small subset of all possible facts. Link prediction is a task of inferring missing facts based on existing ones. We propose TuckER, a relatively straightforward but powerful linear model based on Tucker decomposition of the binary tensor representation of knowledge graph triples. TuckER outperforms previous state-of-the-art models across standard link prediction datasets, acting as a strong baseline for more elaborate models. We show that TuckER is a fully expressive model, derive sufficient bounds on its embedding dimensionalities and demonstrate that several previously introduced linear models can be viewed as special cases of TuckER.
\end{abstract}

\section{Introduction}
\label{introduction}

Vast amounts of information available in the world can be represented succinctly as \emph{entities} and \emph{relations} between them. \emph{Knowledge graphs} are large, graph-structured databases which store facts in triple form $(e_s, r, e_o)$, with $e_s$ and $e_o$ representing subject and object entities and $r$ a relation. However, far from all available information is currently stored in existing knowledge graphs and manually adding new information is costly, which creates the need for algorithms that are able to automatically infer missing facts. 

Knowledge graphs can be represented as a third-order binary tensor, where each element corresponds to a triple, 1 indicating a true fact and 0 indicating the unknown (either a false or a missing fact). The task of \emph{link prediction} is to predict whether two entities are related, based on known facts already present in a knowledge graph, i.e. to infer which of the 0 entries in the tensor are indeed false, and which are missing but actually true.

A large number of approaches to link prediction so far have been linear, based on various methods of factorizing the third-order binary tensor \cite{nickel2011three, yang2014embedding, trouillon2016complex, kazemi2018simple}. Recently, state-of-the-art results have been achieved using non-linear convolutional models \cite{dettmers2018convolutional, balazevic2019hypernetwork}. Despite achieving very good performance, the fundamental problem with deep, non-linear models is that they are non-transparent and poorly understood, as opposed to more mathematically principled and widely studied tensor decomposition models.

In this paper, we introduce TuckER (E stands for entities, R for relations), a straightforward linear model for link prediction on knowledge graphs, based on \emph{Tucker decomposition} \cite{tucker1966some} of the binary tensor of triples, acting as a strong baseline for more elaborate models. Tucker decomposition, used widely in machine learning \cite{schein2016bayesian, ben2017mutan, yang2017deep}, factorizes a tensor into a core tensor multiplied by a matrix along each mode. It can be thought of as a form of higher-order SVD in the special case where matrices are orthogonal and the core tensor is ``all-orthogonal" \cite{ kroonenberg1980principal}. In our case, rows of the matrices contain entity and relation embeddings, while entries of the core tensor determine the level of interaction between them. Subject and object entity embedding matrices are assumed equivalent, i.e. we make no distinction between the embeddings of an entity depending on whether it appears as a subject or as an object in a particular triple. Due to the low rank of the core tensor, TuckER benefits from \textit{multi-task learning} by parameter sharing across relations.

A link prediction model should have enough expressive power to represent all relation types (e.g. symmetric, asymmetric, transitive). We thus show that TuckER is \textit{fully expressive}, i.e. given any ground truth over the triples, there exists an assignment of values to the entity and relation embeddings that accurately separates the true triples from false ones. We also derive a dimensionality bound which guarantees full expressiveness. 

Finally, we show that several previous state-of-the-art linear models, RESCAL \cite{nickel2011three}, DistMult \cite{yang2014embedding}, ComplEx \cite{trouillon2016complex} and SimplE \cite{kazemi2018simple}, are special cases of TuckER.

In summary, key contributions of this paper are:
\begin{itemize}
    \itemsep-0.2em 
    \item proposing TuckER, a new \emph{linear} model for link prediction on knowledge graphs, that is simple, expressive and achieves \emph{state-of-the-art results} across all standard datasets;
    \item proving that TuckER is \emph{fully expressive} and deriving a bound on the embedding dimensionality for full expressiveness; and
    \item showing how TuckER subsumes several previously proposed tensor factorization approaches to link prediction.
\end{itemize}

\section{Related Work} \label{relwork}

\begin{table*}[!t]
\centering
\resizebox{14.5cm}{!}{
\begin{tabular}{lcccc}
  \toprule
  Model & Scoring Function & Relation Parameters & Space Complexity\\
  \midrule
   RESCAL \cite{nickel2011three} & $\mathbf{e}_s^\top\mathbf{W}_r\mathbf{e}_o$ & $\mathbf{W}_r \in \mathbb{R}^{{d_e}^2}$ & $\mathcal{O}(n_e d_e + n_r d_r^2)$ \\
   DistMult \cite{yang2014embedding} & $\langle\mathbf{e}_s, \mathbf{w}_r, \mathbf{e}_o \rangle$ & $\mathbf{w}_r \in \mathbb{R}^{d_e}$ & $\mathcal{O}(n_e d_e + n_r d_e)$ \\
   ComplEx \cite{trouillon2016complex} & $\text{Re}(\langle\mathbf{e}_s, \mathbf{w}_r, \overline{\mathbf{e}}_o \rangle)$ & $\mathbf{w}_r \in \mathbb{C}^{d_e}$ & $\mathcal{O}(n_e d_e + n_r d_e)$ \\
   ConvE \cite{dettmers2018convolutional} & $f(\text{vec}(f([\underline{\mathbf{e}}_s; \underline{\mathbf{w}}_r] * w))\mathbf{W})\mathbf{e}_o$ & $\mathbf{w}_r \in \mathbb{R}^{d_r}$ & $\mathcal{O}(n_e d_e + n_r d_r)$\\
   SimplE \cite{kazemi2018simple} & $\frac{1}{2}(\langle\mathbf{h}_{e_s}, \mathbf{w}_r, \mathbf{t}_{e_o} \rangle + \langle\mathbf{h}_{e_o}, \mathbf{w}_{r^{-1}}, \mathbf{t}_{e_s} \rangle)$ & $\mathbf{w}_r \in \mathbb{R}^{d_e}$ & $\mathcal{O}(n_e d_e + n_r d_e)$ \\
   HypER \cite{balazevic2019hypernetwork} & $f(\text{vec}(\mathbf{e}_s * \text{vec}^{-1}(\mathbf{w}_r\mathbf{H}))\mathbf{W})\mathbf{e}_o$ & $\mathbf{w}_r \in \mathbb{R}^{d_r}$ & $\mathcal{O}(n_e d_e + n_r d_r)$\\
   TuckER (ours) & $\mathcal{W} \times_1 \mathbf{e}_s \times_2 \mathbf{w}_r \times_3 \mathbf{e}_o$ & $\mathbf{w}_r \in \mathbb{R}^{d_r}$ & $\mathcal{O}(n_e d_e + n_r d_r)$\\
  \bottomrule
\end{tabular}
}
  \caption{Scoring functions of state-of-the-art link prediction models, the dimensionality of their relation parameters, and significant terms of their space complexity. $d_e$ and $d_r$ are the dimensionalities of entity and relation embeddings, while $n_e$ and $n_r$ denote the number of entities and relations respectively. $\overline{\mathbf{e}}_o \in \mathbb{C}^{d_e}$ is the complex conjugate of $\mathbf{e}_o$, $\underline{\mathbf{e}}_s, \underline{\mathbf{w}}_r \in \mathbb{R}^{d_w\times d_h}$ denote a 2D reshaping of $\mathbf{e}_s$ and $\mathbf{w}_r$ respectively, $\mathbf{h}_{e_s}, \mathbf{t}_{e_s} \in \mathbb{R}^{d_e}$ are the head and tail entity embedding of entity $e_s$, and $\mathbf{w}_{r^{-1}} \in \mathbb{R}^{d_r}$ is the embedding of relation $r^{-1}$ (which is the inverse of relation $r$). $*$ is the convolution operator, $\langle\cdot\rangle$ denotes the dot product and $\times_n$ denotes the tensor product along the $n$-th mode, $f$ is a non-linear function, and $\mathcal{W} \in \mathbb{R}^{d_e \times d_e \times d_r}$ is the core tensor of a Tucker decomposition.}
  \label{table:models}
\end{table*}

Several \textit{linear} models for link prediction have previously been proposed:

\noindent\textbf{RESCAL} \cite{nickel2011three} optimizes a scoring function containing a bilinear product between subject and object entity vectors and a full rank relation matrix. Although a very expressive and powerful model, RESCAL is prone to overfitting due to its large number of parameters, which increases quadratically in the embedding dimension with the number of relations in a knowledge graph. 

\noindent\textbf{DistMult} \cite{yang2014embedding} is a special case of RESCAL with a diagonal matrix per relation, which reduces overfitting. However, the linear transformation performed on entity embedding vectors in DistMult is limited to a stretch. The binary tensor learned by DistMult is symmetric in the subject and object entity mode and thus DistMult cannot model asymmetric relations. 

\noindent\textbf{ComplEx} \cite{trouillon2016complex} extends DistMult to the complex domain. Subject and object entity embeddings for the same entity are complex conjugates, which introduces asymmetry into the tensor decomposition and thus enables ComplEx to model asymmetric relations.

\noindent\textbf{SimplE} \cite{kazemi2018simple} is based on Canonical Polyadic (CP) decomposition \cite{hitchcock1927expression}, in which subject and object entity embeddings for the same entity are independent (note that DistMult is a special case of CP). SimplE's scoring function alters CP to make subject and object entity embedding vectors dependent on each other by computing the average of two terms, first of which is a bilinear product of the subject entity head embedding, relation embedding and object entity tail embedding and the second is a bilinear product of the object entity head embedding, inverse relation embedding and subject entity tail embedding.

Recently, state-of-the-art results have been achieved with \textit{non-linear} models:

\noindent\textbf{ConvE} \cite{dettmers2018convolutional} performs a global 2D convolution operation on the subject entity and relation embedding vectors, after they are reshaped to matrices and concatenated. The obtained feature maps are flattened, transformed through a linear layer, and the inner product is taken with all object entity vectors to generate a score for each triple. Whilst results achieved by ConvE are impressive, its reshaping and concatenating of vectors as well as using 2D convolution on word embeddings is unintuitive. 

\noindent\textbf{HypER} \cite{balazevic2019hypernetwork} is a simplified convolutional model, that uses a hypernetwork to generate 1D convolutional filters for each relation, extracting relation-specific features from subject entity embeddings. The authors show that convolution is a way of introducing sparsity and parameter tying and that HypER can be understood in terms of tensor factorization up to a non-linearity, thus placing  HypER closer to the well established family of factorization models. The drawback of HypER is that it sets most elements of the core weight tensor to 0, which amounts to hard regularization, rather than letting the model learn which parameters to use via soft regularization. 

Scoring functions of all models described above and TuckER are summarized in Table~\ref{table:models}.

\section{Background}

Let $\mathcal{E}$ denote the set of all entities and $\mathcal{R}$ the set of all relations present in a knowledge graph. A triple is represented as $(e_s, r, e_o)$, with $e_s, e_o \in \mathcal{E}$ denoting subject and object entities respectively and $r \in \mathcal{R}$ the relation between them. 

\subsection{Link Prediction}

In link prediction, we are given a subset of all true triples and the aim is to learn a \textit{scoring function} $\phi$ that assigns a score $s = \phi(e_s, r, e_o) \in \mathbb{R}$ which indicates whether a triple is true, with the ultimate goal of being able to correctly score all missing triples. The scoring function is either a specific form of tensor factorization in the case of linear models or a more complex (deep) neural network architecture for non-linear models. Typically, a positive score for a particular triple indicates a true fact predicted by the model, while a negative score indicates a false one. With most recent models, a non-linearity such as the logistic sigmoid function is typically applied to the score to give a corresponding probability prediction $p = \sigma(s) \in [0, 1]$ as to whether a certain fact is true.

 \subsection{Tucker Decomposition} \label{tuckerdecomp}
 
Tucker decomposition, named after Ledyard R. Tucker \cite{tucker1964extension}, decomposes a tensor into a set of matrices and a smaller core tensor. In a three-mode case, given the original tensor $\mathcal{X} \in \mathbb{R}^{I \times J\times K}$, Tucker decomposition outputs a tensor $\mathcal{Z} \in \mathbb{R}^{P \times Q\times R}$ and three matrices $\mathbf{A} \in \mathbb{R}^{I \times P}$, $\mathbf{B} \in \mathbb{R}^{J \times Q}$, $\mathbf{C} \in \mathbb{R}^{K \times R}$:
\begin{equation}
\mathcal{X} \approx \mathcal{Z} \times_1 \mathbf{A} \times_2 \mathbf{B} \times_3 \mathbf{C},
\label{eq_tucker}
\end{equation}
with $\times_n$ indicating the tensor product along the n-th mode. \textit{Factor matrices} $\mathbf{A}$, $\mathbf{B}$ and $\mathbf{C}$, when orthogonal, can be thought of as the principal components in each mode. Elements of the \textit{core tensor} $\mathcal{Z}$ show the level of interaction between the different components. Typically, $P$, $Q$, $R$ are smaller than $I$, $J$, $K$ respectively, so $\mathcal{Z}$ can be thought of as a compressed version of $\mathcal{X}$. 
Tucker decomposition is not unique, i.e. we can transform $\mathcal{Z}$ without affecting the fit if we apply the inverse transformation to $\mathbf{A}$, $\mathbf{B}$ and $\mathbf{C}$ \cite{kolda2009tensor}. 
 
 \section{Tucker Decomposition for Link Prediction} \label{sec:tucker}

We propose a model that uses Tucker decomposition for link prediction on the binary tensor representation of a knowledge graph, with entity embedding matrix $\mathbf{E}$ that is equivalent for subject and object entities, i.e. $\mathbf{E} = \mathbf{A} = \mathbf{C} \in \mathbb{R}^{n_e \times d_e}$ and relation embedding matrix $\mathbf{R} = \mathbf{B} \in \mathbb{R}^{n_r \times d_r}$, where $n_e$ and $n_r$ represent the number of entities and relations and $d_e$ and $d_r$ the dimensionality of entity and relation embedding vectors. 
\vspace{-0.3cm}
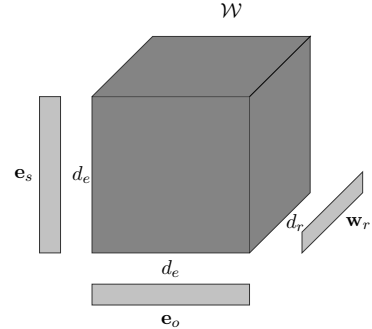
\begin{figure}[H]
\centering
\resizebox{0.65\linewidth}{!}{
\begin{tikzpicture}

\node[black] at (2.5, 0.5, -3) {$\mathcal{W}$};
\draw[black,opacity=0.8,fill=black!60] (1, 0, 0) -- (1, -3, 0) -- (4, -3, 0) -- (4, 0, 0) -- cycle;
\draw[black,opacity=0.8,fill=black!60] (4, 0, 0) -- (4, 0, -3) -- (4, -3, -3) -- (4, -3, 0);
\draw[black,opacity=0.8,fill=black!60] (1, 0, 0) -- (1, 0, -3) -- (4, 0, -3) -- (4, 0, 0);
\node[black] at (2.5, -3.3, 0) {$d_e$};
\node[black] at (0.8, -1.5, 0) {$d_e$};
\node[black] at (4.3, -3, -1.5) {$d_r$};

\node[black] at (-0.3, -1.5, 0) {$\mathbf{e}_s$};
\draw[black,opacity=0.8,fill=black!30] (0, 0, 0) -- (0.4, 0, 0) -- (0.4, -3, 0) -- (0, -3, 0) -- cycle;

\node[black] at (2.5, -4.3, 0) {$\mathbf{e}_o$};
\draw[black,opacity=0.8,fill=black!30] (1, -4, 0) -- (4, -4, 0) -- (4, -3.6, 0) -- (1, -3.6, 0) -- cycle;

\node[black] at (5.5, -3, -1.5) {$\mathbf{w}_r$};
\draw[black,opacity=0.8,fill=black!30] (5, -3, 0) -- (5, -2.6, 0) -- (5, -2.6, -3) -- (5, -3, -3) -- cycle;

\end{tikzpicture}
}
\vspace{-0.2cm}
\caption{Visualization of the TuckER architecture.}
\label{fig:cube}
\end{figure}
\vspace{-0.2cm}
We define the scoring function for TuckER as:
\begin{equation}
\phi(e_s, r, e_o) = \mathcal{W} \times_1 \mathbf{e}_s \times_2 \mathbf{w}_r \times_3 \mathbf{e}_o,
\end{equation}
where $\mathbf{e}_s, \mathbf{e}_o \in \mathbb{R}^{d_e}$ are the rows of $\mathbf{E}$ representing the subject and object entity embedding vectors, $\mathbf{w}_r \in \mathbb{R}^{d_r}$ the rows of $\mathbf{R}$ representing the relation embedding vector and $\mathcal{W} \in \mathbb{R}^{d_e \times d_r \times d_e}$ is the core tensor. We apply logistic sigmoid to each score $\phi(e_s, r, e_o)$ to obtain the predicted probability $p$ of a triple being true. Visualization of the TuckER architecture can be seen in Figure \ref{fig:cube}. As proven in Section \ref{sec:expressive}, TuckER is \textit{fully expressive}. Further, its number of parameters increases \emph{linearly} with respect to entity and relation embedding dimensionality $d_e$ and $d_r$, as the number of entities and relations increases, since the number of parameters of $\mathcal{W}$ depends only on the entity and relation embedding dimensionality and not on the number of entities or relations. By having the core tensor $\mathcal{W}$, unlike simpler models such as DistMult, ComplEx and SimplE, TuckER does not encode all the learned knowledge into the embeddings; some is stored in the core tensor and shared between all entities and relations through \textit{multi-task learning}. Rather than learning distinct relation-specific matrices, the core tensor of TuckER can be viewed as containing a shared pool of ``prototype'' relation matrices, which are linearly combined according to the parameters in each relation embedding. 

\subsection{Training}

Since the logistic sigmoid is applied to the scoring function to approximate the true binary tensor, the implicit underlying tensor is comprised of $-\infty$ and $\infty$. Given this prevents an explicit analytical factorization, we use numerical methods to train TuckER. We use the standard data augmentation technique, first used by \citet{dettmers2018convolutional} and formally described by \citet{lacroix2018canonical}, of adding reciprocal relations for every triple in the dataset, i.e. we add $(e_o, r^{-1}, e_s)$ for every $(e_s, r, e_o)$. Following the training procedure introduced by \citet{dettmers2018convolutional} to speed up training, we use \textit{1-N scoring}, i.e. we simultaneously score entity-relation pairs $(e_s, r)$ and $(e_o, r^{-1})$ with all entities $e_o \in \mathcal{E}$ and $e_s \in \mathcal{E}$ respectively, in contrast to \textit{1-1 scoring}, where individual triples $(e_s, r, e_o)$ and $(e_o, r^{-1}, e_s)$ are trained one at a time. The model is trained to minimize the Bernoulli negative log-likelihood loss function. A component of the loss for one entity-relation pair with all others entities is defined as:
\begin{equation}
\resizebox{\hsize}{!}{
$L = -\frac{1}{n_e}\sum\limits_{i=1}^{n_e} (\mathbf{y}^{(i)} \text{log}(\mathbf{p}^{(i)}) + (1-\mathbf{y}^{(i)}) \text{log}(1-\mathbf{p}^{(i)})),$}
\end{equation}
where $\mathbf{p} \in \mathbb{R}^{n_e}$ is the vector of predicted probabilities and $\mathbf{y} \in \mathbb{R}^{n_e}$ is the binary label vector.

\section{Theoretical Analysis}
\subsection{Full Expressiveness and Embedding Dimensionality} \label{sec:expressive}

A tensor factorization model is fully expressive if for any ground truth over all entities and relations, there exist entity and relation embeddings that accurately separate true triples from the false. As shown in \cite{trouillon2017knowledge}, ComplEx is fully expressive with the embedding dimensionality bound $d_e=d_r=n_e \cdot n_r$. Similarly to ComplEx, \citet{kazemi2018simple} show that SimplE is fully expressive with entity and relation embeddings of size $d_e=d_r=\textit{min}(n_e \cdot n_r, \gamma + 1)$, where $\gamma$ represents the number of true facts. They further prove other models are not fully expressive: DistMult, because it cannot model asymmetric relations; and transitive models such as TransE \cite{bordes2013translating} and its variants FTransE \cite{feng2016knowledge} and STransE \cite{nguyen2016stranse}, because of certain contradictions that they impose between different relation types. By Theorem \ref{theorem1}, we establish the bound on entity and relation embedding dimensionality (i.e. decomposition rank) that guarantees full expressiveness of TuckER.

\begin{theorem} \label{theorem1}
Given any ground truth over a set of entities $\mathcal{E}$ and relations $\mathcal{R}$, there exists a TuckER model with entity embeddings of dimensionality $d_e=n_e$ and relation embeddings of dimensionality $d_r=n_r$, where $n_e=|\mathcal{E}|$ is the number of entities and $n_r = |\mathcal{R}|$ the number of relations, that accurately represents that ground truth.
\end{theorem} 
\vspace{-0.2cm}
\begin{proof}
Let $\mathbf{e}_s$ and $\mathbf{e}_o$ be the $n_e$-dimensional one-hot binary vector representations of subject and object entities $e_s$ and $e_o$ respectively and $\mathbf{w}_r$ the $n_r$-dimensional one-hot binary vector representation of relation $r$. For each subject entity $e_s^{(i)}$, relation $r^{(j)}$ and object entity $e_o^{(k)}$, we let the $i$-th, $j$-th and $k$-th element respectively of the corresponding vectors $\mathbf{e}_s$, $\mathbf{w}_r$ and $\mathbf{e}_o$ be 1 and all other elements 0. Further, we set the $ijk$ element of the tensor $\mathcal{W} \in \mathbb{R}^{n_e \times n_r \times n_e}$ to 1 if the fact $(e_s, r, e_o)$ holds and -1 otherwise. Thus the product of the entity embeddings and the relation embedding with the core tensor, after applying the logistic sigmoid, accurately represents the original tensor.
\end{proof}

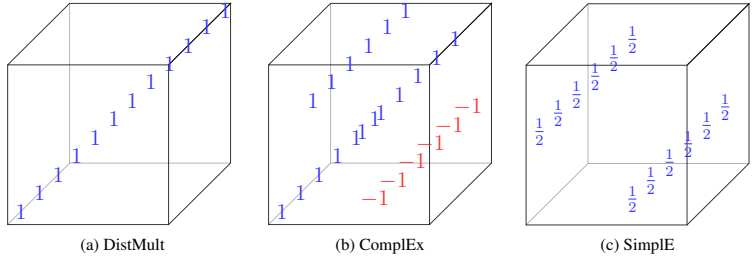
\begin{figure*}[!ht]
\centering
\resizebox{10.3cm}{!}{

\begin{subfigure}[b]{0.3\textwidth}
    \centering
    \begin{tikzpicture}
    \draw[black,opacity=0.3] (1, -3, 0) -- (1, -3, -3) -- (4, -3, -3);
    \draw[black,opacity=0.3] (1, -3, -3) -- (1, 0, -3);
    \draw[black,opacity=0.8] (1, 0, 0) -- (1, -3, 0) -- (4, -3, 0) -- (4, 0, 0) -- cycle;
    \draw[black,opacity=0.8] (4, 0, 0) -- (4, 0, -3) -- (4, -3, -3) -- (4, -3, 0);
    \draw[black,opacity=0.8] (1, 0, 0) -- (1, 0, -3) -- (4, 0, -3) -- (4, 0, 0);
    
    \node[white!20!blue] at (1.25, -2.75, 0) {\large$1$};
    \node[white!20!blue] at (1.5, -2.5, -0.25) {\large$1$};
    \node[white!20!blue] at (1.75, -2.25, -0.5) {\large$1$};
    \node[white!20!blue] at (2, -2, -0.75) {\large$1$};
    \node[white!20!blue] at (2.25, -1.75, -1) {\large$1$};
    \node[white!20!blue] at (2.5, -1.5, -1.25) {\large$1$};
    \node[white!20!blue] at (2.75, -1.25, -1.5) {\large$1$};
    \node[white!20!blue] at (3, -1, -1.75) {\large$1$};
    \node[white!20!blue] at (3.25, -0.75, -2) {\large$1$};
    \node[white!20!blue] at (3.5, -0.5, -2.25) {\large$1$};
    \node[white!20!blue] at (3.75, -0.25, -2.5) {\large$1$};
    \node[white!20!blue] at (4, -0.05, -2.75) {\large$1$};
    \end{tikzpicture}
    \caption{DistMult}
    \label{fig:distmult}
    
\end{subfigure}\hfill
\begin{subfigure}[b]{0.3\textwidth}
    \centering
    \begin{tikzpicture}
    \draw[black,opacity=0.3] (1, -3, 0) -- (1, -3, -3) -- (4, -3, -3);
    \draw[black,opacity=0.3] (1, -3, -3) -- (1, 0, -3);
    \draw[black,opacity=0.8] (1, 0, 0) -- (1, -3, 0) -- (4, -3, 0) -- (4, 0, 0) -- cycle;
    \draw[black,opacity=0.8] (4, 0, 0) -- (4, 0, -3) -- (4, -3, -3) -- (4, -3, 0);
    \draw[black,opacity=0.8] (1, 0, 0) -- (1, 0, -3) -- (4, 0, -3) -- (4, 0, 0);
    \node[white!20!blue] at (1.25, -2.75, 0) {\large$1$};
    \node[white!20!blue] at (1.5, -2.5, -0.25) {\large$1$};
    \node[white!20!blue] at (1.75, -2.25, -0.5) {\large$1$};
    \node[white!20!blue] at (2, -2, -0.75) {\large$1$};
    \node[white!20!blue] at (2.25, -1.75, -1) {\large$1$};
    \node[white!20!blue] at (2.5, -1.5, -1.25) {\large$1$};
    
    \node[white!20!blue] at (2.75, -1.25, 0) {\large$1$};
    \node[white!20!blue] at (3, -1., -0.25) {\large$1$};
    \node[white!20!blue] at (3.25, -0.75, -0.5) {\large$1$};
    \node[white!20!blue] at (3.5, -0.5, -0.75) {\large$1$};
    \node[white!20!blue] at (3.75, -0.25, -1) {\large$1$};
    \node[white!20!blue] at (4, 0, -1.25) {\large$1$};
    
    \node[white!20!blue] at (1.25, -1.25, -1.5) {\large$1$};
    \node[white!20!blue] at (1.5, -1, -1.75) {\large$1$};
    \node[white!20!blue] at (1.75, -0.75, -2) {\large$1$};
    \node[white!20!blue] at (2, -0.5, -2.25) {\large$1$};
    \node[white!20!blue] at (2.25, -0.25, -2.5) {\large$1$};
    \node[white!20!blue] at (2.5, -0.05, -2.75) {\large$1$};
    
    \node[white!20!red] at (2.5, -3, -1.25) {\large$-1$};
    \node[white!20!red] at (2.75, -2.75, -1.5) {\large$-1$};
    \node[white!20!red] at (3, -2.5, -1.75) {\large$-1$};
    \node[white!20!red] at (3.25, -2.25, -2) {\large$-1$};
    \node[white!20!red] at (3.5, -2, -2.25) {\large$-1$};
    \node[white!20!red] at (3.75, -1.75, -2.5) {\large$-1$};
    \end{tikzpicture}
    \caption{ComplEx}
    \label{fig:complex}
\end{subfigure}\hfill
\begin{subfigure}[b]{0.3\textwidth}
    \centering
    \begin{tikzpicture}
    \draw[black,opacity=0.3] (1, -3, 0) -- (1, -3, -3) -- (4, -3, -3);
    \draw[black,opacity=0.3] (1, -3, -3) -- (1, 0, -3);
    \draw[black,opacity=0.8] (1, 0, 0) -- (1, -3, 0) -- (4, -3, 0) -- (4, 0, 0) -- cycle;
    \draw[black,opacity=0.8] (4, 0, 0) -- (4, 0, -3) -- (4, -3, -3) -- (4, -3, 0);
    \draw[black,opacity=0.8] (1, 0, 0) -- (1, 0, -3) -- (4, 0, -3) -- (4, 0, 0);
    
    \node[white!20!blue] at (1.25, -1.25, 0) {\large$\frac{1}{2}$};
    \node[white!20!blue] at (1.5, -1, -0.25) {\large$\frac{1}{2}$};
    \node[white!20!blue] at (1.75, -0.75, -0.5) {\large$\frac{1}{2}$};
    \node[white!20!blue] at (2, -0.5, -0.75) {\large$\frac{1}{2}$};
    \node[white!20!blue] at (2.25, -0.25, -1) {\large$\frac{1}{2}$};
    \node[white!20!blue] at (2.5, 0, -1.25) {\large$\frac{1}{2}$};
    
    \node[white!20!blue] at (2.5, -3, -1.25) {\large$\frac{1}{2}$};
    \node[white!20!blue] at (2.75, -2.75, -1.5) {\large$\frac{1}{2}$};
    \node[white!20!blue] at (3, -2.5, -1.75) {\large$\frac{1}{2}$};
    \node[white!20!blue] at (3.25, -2.25, -2) {\large$\frac{1}{2}$};
    \node[white!20!blue] at (3.5, -2, -2.25) {\large$\frac{1}{2}$};
    \node[white!20!blue] at (3.75, -1.75, -2.5) {\large$\frac{1}{2}$};
    
    \end{tikzpicture}
    \caption{SimplE}
    \label{fig:simple}
\end{subfigure}
}
\caption{Constraints imposed on the values of core tensor $\mathcal{Z} \in \mathbb{R}^{d_e \times d_e \times d_e}$ for DistMult and $\mathcal{Z} \in \mathbb{R}^{2d_e \times 2d_e \times 2d_e}$ for ComplEx and SimplE. Elements that are set to 0 are represented in white.}
\label{fig:cubes}
\end{figure*}

The purpose of Theorem \ref{theorem1} is to prove that TuckER is capable of potentially capturing all information (and noise) in the data. In practice however, we expect the embedding dimensionalities needed for full reconstruction of the underlying binary tensor to be much smaller than the bound stated above, since the assignment of values to the tensor is not random but follows a certain structure, otherwise nothing unknown could be predicted. Even more so, low decomposition rank is actually a desired property of any bilinear link prediction model, forcing it to learn that structure and generalize to new data, rather than simply memorizing the input. In general, we expect TuckER to perform better than ComplEx and SimplE with embeddings of lower dimensionality due to parameter sharing in the core tensor (shown empirically in Section \ref{sec:bound}), which could be of importance for efficiency in downstream tasks.

\subsection{Relation to Previous Linear Models}

Several previous tensor factorization models can be viewed as a special case of TuckER:

\keypoint{RESCAL \cite{nickel2011three}} Following the notation introduced in Section \ref{tuckerdecomp}, the RESCAL scoring function (see Table \ref{table:models}) has the form: 
\vspace{-0.2cm}
\begin{equation}
\mathcal{X} \approx \mathcal{Z} \times_1 \mathbf{A} \times_3 \mathbf{C}.
\label{eq_tucker2}
\vspace{-0.2cm}
\end{equation}
This corresponds to Equation \ref{eq_tucker} with $I = K = n_e$, $P = R = d_e$, $Q = J = n_r$ and $\mathbf{B} = \mathbf{I}_J$ the $J \times J$ identity matrix. This is also known as Tucker2 decomposition \cite{kolda2009tensor}. As is the case with TuckER, the entity embedding matrix of RESCAL is shared between subject and object entities, i.e. $\mathbf{E} = \mathbf{A} = \mathbf{C} \in \mathbb{R}^{n_e \times d_e}$ and the relation matrices $\mathbf{W}_r \in \mathbb{R}^{d_e \times d_e}$ are the $d_e \times d_e$ slices of the core tensor $\mathcal{Z}$. As mentioned in Section \ref{relwork}, the drawback of RESCAL compared to TuckER is that its number of parameters grows \emph{quadratically} in the entity embedding dimension $d_e$ as the number of relations increases. 

\keypoint{DistMult \cite{yang2014embedding}} The scoring function of DistMult (see Table \ref{table:models}) can be viewed as  equivalent to that of TuckER (see Equation \ref{eq_tucker}) with a core tensor $\mathcal{Z} \in \mathbb{R}^{P \times Q \times R}$, $P = Q = R = d_e$, which is \textit{superdiagonal} with 1s on the superdiagonal, i.e. all elements $z_{pqr}$ with $p=q=r$ are 1 and all the other elements are 0 (as shown in Figure \ref{fig:distmult}). Rows of $\mathbf{E} = \mathbf{A} = \mathbf{C} \in \mathbb{R}^{n_e \times d_e}$ contain subject and object entity embedding vectors $\mathbf{e}_s, \mathbf{e}_o \in \mathbb{R}^{d_e}$ and rows of $\mathbf{R} = \mathbf{B} \in \mathbb{R}^{n_r \times d_e}$ contain relation embedding vectors $\mathbf{w}_r \in \mathbb{R}^{d_e}$. It is interesting to note that the TuckER interpretation of the DistMult scoring function, given that matrices $\mathbf{A}$ and $\mathbf{C}$ are identical, can alternatively be interpreted as a special case of CP decomposition \cite{hitchcock1927expression}, since Tucker decomposition with a superdiagonal core tensor is equivalent to CP decomposition. Due to enforced symmetry in subject and object entity mode, DistMult cannot learn to represent asymmetric relations.

\keypoint{ComplEx \cite{trouillon2016complex}} \textit{Bilinear models} represent subject and object entity embeddings as vectors $\mathbf{e}_s, \mathbf{e}_o \in \mathbb{R}^{d_e}$, relation as a matrix $\mathbf{W}_r \in \mathbb{R}^{d_e \times d_e}$ and the scoring function as a bilinear product $\phi(e_s, r, e_o) = \mathbf{e}_s\mathbf{W}_r\mathbf{e}_o$. It is trivial to show that both RESCAL and DistMult belong to the family of bilinear models. As explained by \citet{kazemi2018simple}, ComplEx can be considered a bilinear model with the real and imaginary part of an embedding for each entity concatenated in a single vector, $[\text{Re}(\mathbf{e}_s); \text{Im}(\mathbf{e}_s)] \in \mathbb{R}^{2d_e}$ for subject, $[\text{Re}(\mathbf{e}_o); \text{Im}(\mathbf{e}_o)] \in \mathbb{R}^{2d_e}$ for object, and a relation matrix $\mathbf{W}_r \in \mathbb{R}^{2d_e \times 2d_e}$, constrained so that its leading diagonal contains duplicated elements of $\text{Re}(\mathbf{w}_r)$, its $d_e$-diagonal elements of $\text{Im}(\mathbf{w}_r)$ and its -$d_e$-diagonal elements of -$\text{Im}(\mathbf{w}_r)$, with all other elements set to 0, where $d_e$ and -$d_e$ represent offsets from the leading diagonal. 

Similarly to DistMult, we can regard the scoring function of ComplEx (see Table \ref{table:models}) as equivalent to the scoring function of TuckER (see Equation \ref{eq_tucker}), with core tensor $\mathcal{Z} \in \mathbb{R}^{P \times Q \times R}$, $P=Q=R=2d_e$, where $3d_e$ elements on different tensor diagonals are set to 1, $d_e$ elements on one tensor diagonal are set to -1 and all other elements are set to 0 (see Figure \ref{fig:complex}). This shows that the scoring function of ComplEx, which computes a bilinear product with \textit{complex} entity and relation embeddings and disregards the imaginary part of the obtained result, is equivalent to a hard regularization of the core tensor of TuckER in the \textit{real} domain.  

\keypoint{SimplE \cite{kazemi2018simple}} The authors show that SimplE belongs to the family of bilinear models by concatenating embeddings for head and tail entities for both subject and object into vectors $[\mathbf{h}_{e_s};\mathbf{t}_{e_s}] \in \mathbb{R}^{2d_e}$ and $[\mathbf{h}_{e_o};\mathbf{t}_{e_o}] \in \mathbb{R}^{2d_e}$ and constraining the relation matrix $\mathbf{W}_r \in \mathbb{R}^{2d_e \times 2d_e}$ so that it contains the relation embedding vector $\frac{1}{2}\mathbf{w}_r$ on its $d_e$-diagonal and the inverse relation embedding vector $\frac{1}{2}\mathbf{w}_{r^{-1}}$ on its -$d_e$-diagonal and 0s elsewhere. The SimplE scoring function (see Table \ref{table:models}) is therefore equivalent to that of TuckER (see Equation \ref{eq_tucker}), with core tensor $\mathcal{Z} \in \mathbb{R}^{P \times Q \times R}$, $P=Q=R=2d_e$, where $2d_e$ elements on two tensor diagonals are set to $\frac{1}{2}$ and all other elements are set to 0 (see Figure \ref{fig:simple}).

\subsection{Representing Asymmetric Relations}
\begin{figure}[!ht]
\centering
\begin{subfigure}[b]{0.23\textwidth}
    \centering
    \includegraphics[width=0.93\linewidth]{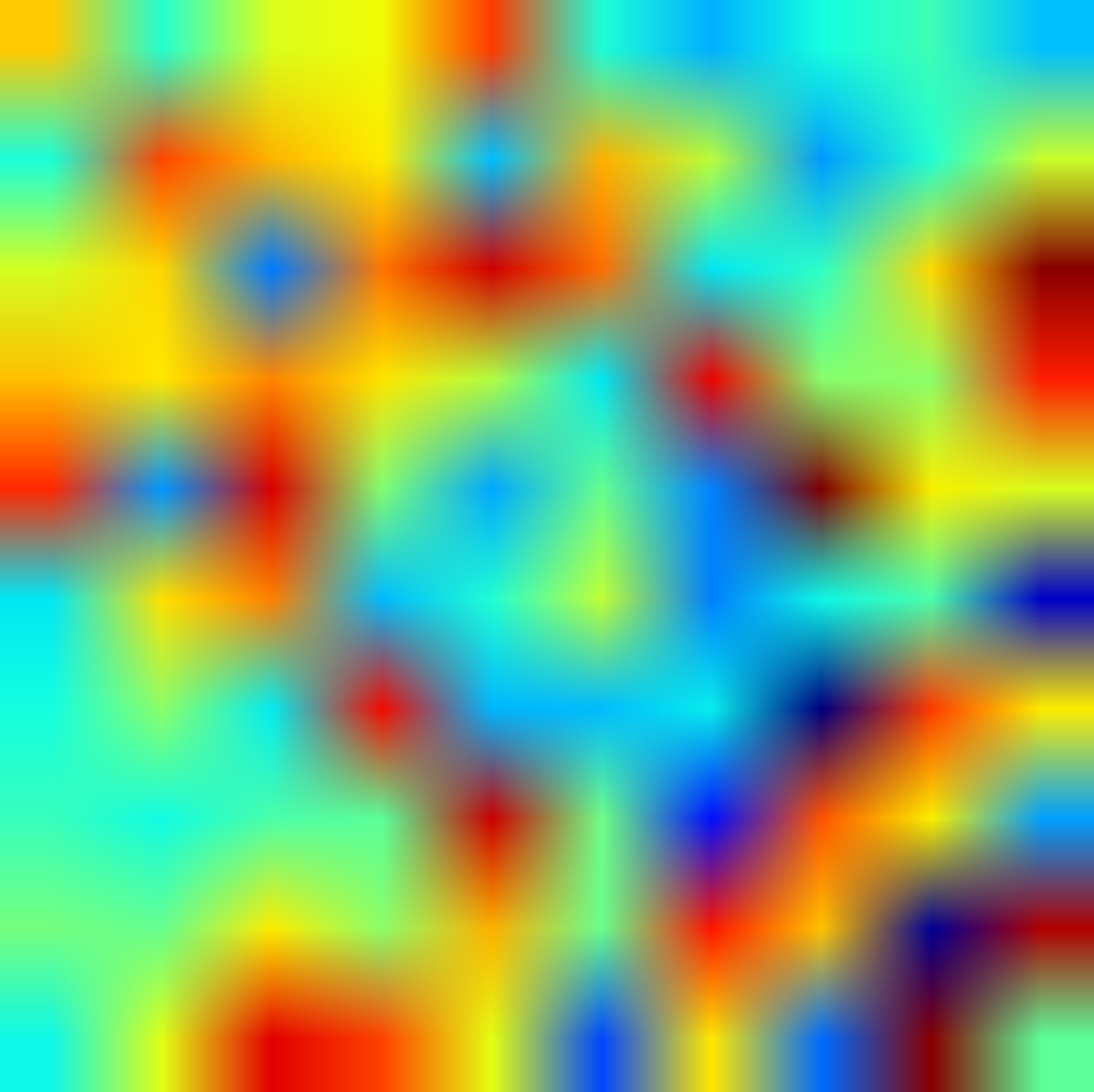}
    \caption{$\mathbf{W}_{\text{derivationally\_related\_form}}$}
    \label{fig:drf}
\end{subfigure}
\begin{subfigure}[b]{0.23\textwidth}
    \centering
    \includegraphics[width=0.93\linewidth]{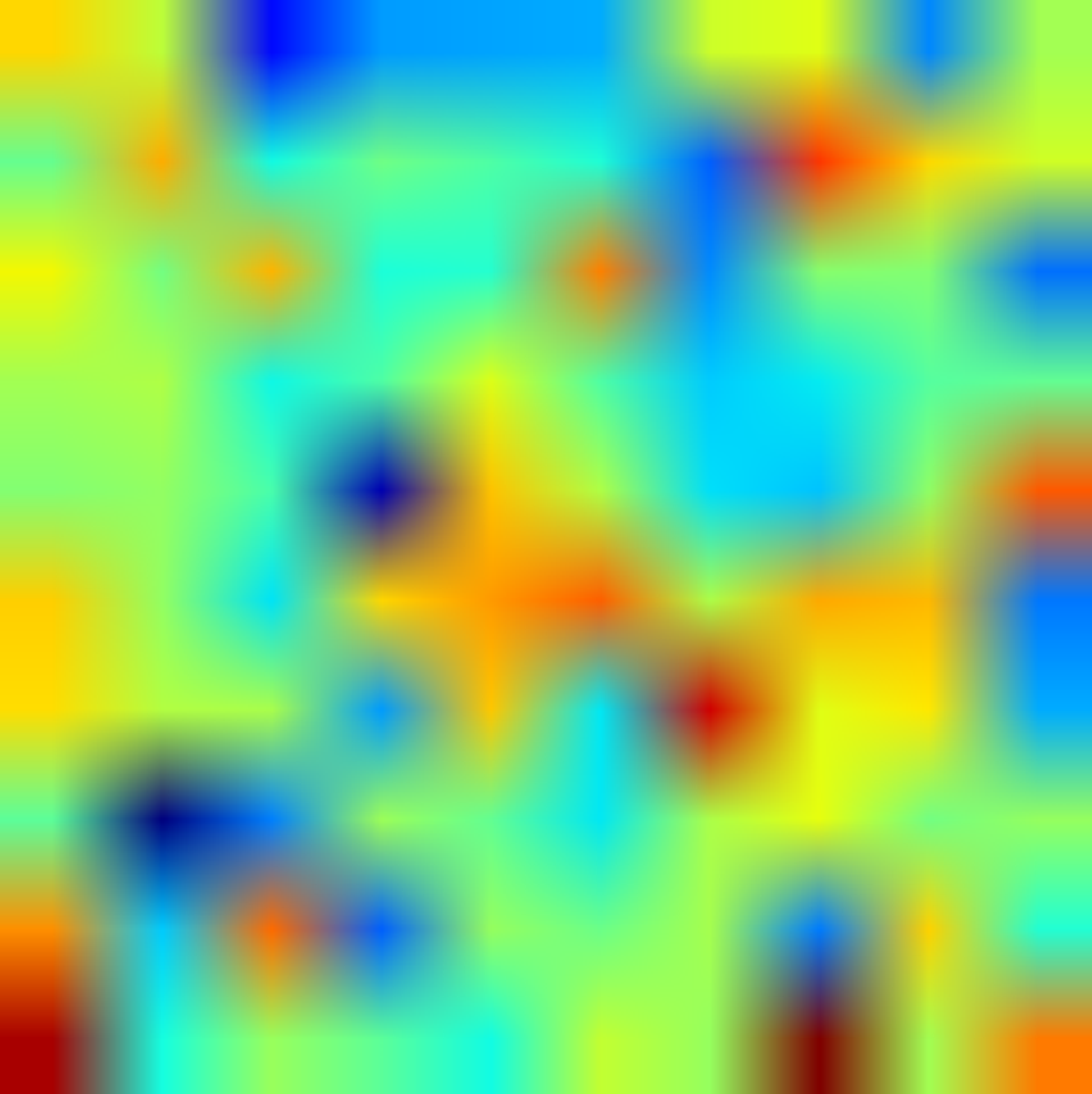}
    \caption{$\mathbf{W}_{\text{hypernym}}$}
    \label{fig:hypernym}
\end{subfigure}
\caption{Learned relation matrices for a symmetric (derivationally\_related\_form) and an asymmetric (hypernym) WN18RR relation. $\mathbf{W}_{\text{derivationally\_related\_form}}$ is approximately symmetric about the leading diagonal.}
\label{fig:asymmetry}
\end{figure}

Each relation in a knowledge graph can be characterized by a certain set of properties, such as symmetry, reflexivity, transitivity. So far, there have been two possible ways in which linear link prediction models introduce \textit{asymmetry} into factorization of the binary tensor of triples:
\vspace{-0.1cm}
\begin{itemize}
    \item distinct (although possibly related) embeddings for subject and object entities and a diagonal matrix (or equivalently a vector) for each relation, as is the case with models such as ComplEx and SimplE; or
    \item equivalent subject and object entity embeddings and each relation represented by a full rank matrix, which is the case with RESCAL.
\end{itemize}
The latter approach appears more intuitive, since asymmetry is a property of the relation, rather than the entities. However, the drawback of the latter approach is quadratic growth of parameter number with the number of relations, which often leads to overfitting, especially for relations with a small number of training triples. TuckER overcomes this by representing relations as vectors $\mathbf{w}_r$, which makes the parameter number grow linearly with the number of relations, while still keeping the desirable property of allowing relations to be asymmetric by having an asymmetric \textit{relation-agnostic} core tensor $\mathcal{W}$, rather than encoding the relation-specific information in the entity embeddings. Multiplying $\mathcal{W} \in \mathbb{R}^{d_e \times d_r \times d_e}$ with $\mathbf{w}_r \in \mathbb{R}^{d_r}$ along the second mode, we obtain a full rank relation-specific matrix $\mathbf{W}_r \in \mathbb{R}^{d_e \times d_e}$, which can perform all possible linear transformations on the entity embeddings, i.e. rotation, reflection or stretch, and is thus also capable of modeling asymmetry. Regardless of what kind of transformation is needed for modeling a particular relation, TuckER can learn it from the data. To demonstrate this, we show sample heatmaps of learned relation matrices $\mathbf{W}_r$ for a WordNet symmetric relation ``derivationally\_related\_form'' and an asymmetric relation ``hypernym'' in Figure \ref{fig:asymmetry}, where one can see that TuckER learns to model the symmetric relation with the relation matrix that is approximately symmetric about the main diagonal, whereas the matrix belonging to the asymmetric relation exhibits no obvious structure.

\vspace{-0.4cm}
\section{Experiments and Results}

\subsection{Datasets}

We evaluate TuckER using four standard link prediction datasets (see Table \ref{datasets}):

\noindent\textbf{FB15k}  \cite{bordes2013translating} is a subset of Freebase, a large database of real world facts.

\noindent\textbf{FB15k-237} \cite{toutanova2015representing} was created from FB15k by removing the inverse of many relations that are present in the training set from validation and test sets, making it more difficult for simple models to do well.

\noindent\textbf{WN18}  \cite{bordes2013translating} is a subset of WordNet, a hierarchical database containing lexical relations between words.

\noindent\textbf{WN18RR} \cite{dettmers2018convolutional} is a subset of WN18, created by removing the inverse relations from validation and test sets. 
 
\subsection{Implementation and Experiments}
We implement TuckER in PyTorch \cite{paszke2017automatic} and make our code available on GitHub.\footnote{\texttt{https://github.com/ibalazevic/TuckER}}

We choose all hyper-parameters by random search based on validation set performance. For FB15k and FB15k-237, we set entity and relation embedding dimensionality to $d_e= d_r = 200$. For WN18 and WN18RR, which both contain a significantly smaller number of relations relative to the number of entities as well as a small number of relations compared to FB15k and FB15k-237, we set $d_e=200$ and $d_r=30$.  We use batch normalization \cite{ioffe2015batch} and dropout \cite{srivastava2014dropout} to speed up training. We find that lower dropout values $(0.1, 0.2)$ are required for datasets with a higher number of training triples per relation and thus less risk of overfitting (WN18 and WN18RR), whereas higher dropout values $(0.3, 0.4, 0.5)$ are required for FB15k and FB15k-237. We choose the learning rate from $\{0.01, 0.005, 0.003, 0.001, 0.0005\}$ and learning rate decay from $\{1, 0.995, 0.99\}$. We find the following combinations of learning rate and learning rate decay to give the best results:  $(0.003, 0.99)$ for FB15k, $(0.0005, 1.0)$ for FB15k-237, $(0.005, 0.995)$ for WN18 and $(0.01, 1.0)$ for WN18RR (see Table \ref{table:hyperparams} in the Appendix \ref{sec:hyperparams} for a complete list of hyper-parameter values on each dataset). We train the model using Adam \cite{kingma2014adam} with the batch size 128.

At evaluation time, for each test triple we generate $n_e$ candidate triples by combining the test entity-relation pair with all possible entities $\mathcal{E}$, ranking the scores obtained. We use the \textit{filtered} setting \cite{bordes2013translating}, i.e. all known true triples are removed from the candidate set except for the current test triple. We use evaluation metrics standard across the link prediction literature: mean reciprocal rank (MRR) and hits@$k$, $k \in \{1, 3, 10\}$. Mean reciprocal rank is the average of the inverse of the mean rank assigned to the true triple over all candidate triples. Hits@$k$ measures the percentage of times a true triple is ranked within the top $k$ candidate triples.

\begin{table}[!htp]
    \centering
    \resizebox{6.2cm}{!}{
      \begin{tabular}{lrr}
        \toprule 
        Dataset & \qquad \# Entities ($n_e$) & \ \ \# Relations ($n_r$)\\
        \midrule 
        FB15k  		&	14,951 	& 1,345 \\
        FB15k-237	&	14,541	& 237	\\
        WN18 		& 	40,943 	& 18 	\\
        WN18RR		& 	40,943 	& 11	\\
        \bottomrule
      \end{tabular}
    }
    \caption{Dataset statistics.}
    \vspace{-0.2cm}
     \label{datasets}
 \end{table}

\begin{table*}[!t]
	\centering
    \resizebox{13cm}{!}{
    \begin{tabular}{lcccccccccc}
    \toprule 
    & &\multicolumn{4}{c}{WN18RR}&&\multicolumn{4}{c}{FB15k-237}\\ 
    \cmidrule{3-6} \cmidrule{8-11}
    & Linear & MRR & Hits@10 & Hits@3 & Hits@1 & & MRR & Hits@10 & Hits@3 & Hits@1\\
    \midrule
	DistMult \cite{yang2014embedding} & yes & $.430$ & $.490$ & $.440$ & $.390$ & & $.241$ & $.419$ & $.263$ & $.155$\\ 
    ComplEx \cite{trouillon2016complex} & yes & $.440$ & $.510$ & $.460$ & $.410$ & & $.247$ & $.428$ & $.275$ & $.158$ \\
    Neural LP \cite{yang2017differentiable} & no & $-$ & $-$ & $-$ & $-$ & & $.250$ & $.408$ & $-$ & $-$ \\
    R-GCN \cite{schlichtkrull2018modeling} & no & $-$ & $-$ & $-$ & $-$ & & $.248$ & $.417$ & $.264$ & $.151$ \\
    MINERVA \cite{das2018go} & no & $-$ & $-$ & $-$ & $-$ & & $-$ & $.456$ & $-$ & $-$\\
    ConvE \cite{dettmers2018convolutional} & no & $.430$ & $.520$ & $.440$ & $.400$ & & $.325$ & $.501$ & $.356$ & $.237$ \\ 
    HypER \cite{balazevic2019hypernetwork} & no & $.465$ & $.522$ & $.477$ & $.436$ & & $.341$ &$.520$& $.376$ & $.252$\\
    M-Walk \cite{shen2018m} & no & $.437$ & $-$ & $.445$ & $.414$ & & $-$ & $-$ & $-$ & $-$ \\ 
    RotatE \cite{sun2019rotate} & no & $-$ & $-$ & $-$ & $-$ & & $.297$ & $.480$ & $.328$ & $.205$\\
    \midrule
    TuckER (ours)  & yes & $\mathbf{.470}$ & $\mathbf{.526}$ & $\mathbf{.482}$ & $\mathbf{.443}$ & & $\mathbf{.358}$ & $\mathbf{.544}$ & $\mathbf{.394}$ & $\mathbf{.266}$ \\ 
    \bottomrule
    \end{tabular}
    }
    \caption{Link prediction results on WN18RR and FB15k-237. The RotatE \cite{sun2019rotate} results are reported without their self-adversarial negative sampling (see Appendix H in the original paper) for fair comparison.}
     \label{table:wn18rr}
 \end{table*}

 \begin{table*}[!t]
    \centering
    \resizebox{13cm}{!}{
    \begin{tabular}{lcccccccccc}
    \toprule 
    & & \multicolumn{4}{c}{WN18}&&\multicolumn{4}{c}{FB15k}\\ 
    \cmidrule{3-6} \cmidrule{8-11}
    & Linear & MRR & Hits@10 & Hits@3 & Hits@1 & & MRR & Hits@10 & Hits@3 & Hits@1\\
    \midrule
    TransE \cite{bordes2013translating} & no & $-$ & $.892$ & $-$ & $-$ & & $-$ & $.471$ & $-$ & $-$\\ 
	DistMult \cite{yang2014embedding} & yes & $.822$ & $.936$ & $.914$ & $.728$ & & $.654$ & $.824$ & $.733$ & $.546$\\ 
    ComplEx \cite{trouillon2016complex} & yes & $.941$ & $.947$ & $.936$ & $.936$ & & $.692$ & $.840$ & $.759$ & $.599$ \\
    ANALOGY \cite{liu2017analogical} & yes & $.942$ & $.947$ & $.944$ & $.939$ & & $.725$ & $.854$ & $.785$ & $.646$ \\
    Neural LP \cite{yang2017differentiable} & no & $.940$ & $.945$ & $-$ & $-$ & & $.760$ & $.837$ & $-$ & $-$ \\
    R-GCN \cite{schlichtkrull2018modeling} & no & $.819$ & $\mathbf{.964}$ & $.929$ & $.697$ & & $.696$ & $.842$ & $.760$ & $.601$ \\
    TorusE \cite{ebisu2018toruse} & no & $.947$ & $.954$ & $.950$ & $.943$ & & $.733$ & $.832$ & $.771$ & $.674$ \\
    ConvE \cite{dettmers2018convolutional} & no & $.943$ & $.956$ & $.946$ & $.935$ & & $.657$ & $.831$ & $.723$ & $.558$ \\ 
    HypER \cite{balazevic2019hypernetwork} & no & $.951$ & $958$ & $\mathbf{.955}$ & $.947$ & & $.790$ & $.885$ & $.829$ & $.734$\\
    SimplE \cite{kazemi2018simple} & yes & $.942$ & $.947$ & $.944$ & $.939$ & & $.727$ & $.838$ & $.773$ & $.660$ \\
    \midrule
    TuckER (ours) & yes & $\mathbf{.953}$ & $.958$ & $\mathbf{.955}$ & $\mathbf{.949}$ & & $\mathbf{.795}$ & $\mathbf{.892}$& $\mathbf{.833}$ & $\mathbf{.741}$\\
    \bottomrule
    \end{tabular}
    }
    \caption{Link prediction results on WN18 and FB15k.}
     \label{table:wn18}
 \end{table*}

\subsection{Link Prediction Results}
Link prediction results on all datasets are shown in Tables \ref{table:wn18rr} and \ref{table:wn18}. Overall, TuckER outperforms previous state-of-the-art models on all metrics across all datasets (apart from hits@10 on WN18 where a non-linear model, R-GCN, does better). Results achieved by TuckER are not only better than those of other linear models, such as DistMult, ComplEx and SimplE, but also better than the results of many more complex deep neural network and reinforcement learning architectures, e.g. R-GCN, MINERVA, ConvE and HypER, demonstrating the expressive power of linear models and supporting our claim that simple linear models should serve as a baseline before moving onto more elaborate models.

Even with fewer parameters than ComplEx and SimplE at $d_e\!=\!200$ and $d_r\!=\!30$ on WN18RR ($\sim$9.4 vs $\sim$16.4 million), TuckER consistently obtains better results than any of those models. We believe this is because TuckER exploits knowledge sharing between relations through the core tensor, i.e. multi-task learning. This is supported by the fact that the margin by which TuckER outperforms other linear models is notably increased on datasets with a large number of relations. For example, improvement on FB15k is $+14\%$ over ComplEx and $+8\%$ over SimplE on the toughest hits@1 metric. To our knowledge, ComplEx-N3 \cite{lacroix2018canonical} is the only other linear link prediction model that benefits from multi-task learning. There, rank regularization of the embedding matrices is used to encourage a low-rank factorization, thus forcing parameter sharing between relations. We do not include their published results in Tables \ref{table:wn18rr} and \ref{table:wn18}, since they use 
the highly non-standard $d_e \!=\! d_r \!=\! 2000$ and thus a far larger parameter number (18x more parameters than TuckER on WN18RR; 5.5x on FB15k-237), making their results incomparable to those typically reported, including our own. However, running their model with equivalent parameter number to TuckER shows comparable performance, supporting our belief that the two models both attain the benefits of multi-task learning, although by different means.

\subsection{Influence of Parameter Sharing} \label{sec:bound}

\begin{figure}[!htb]
\centering
\includegraphics[width=0.75\linewidth]{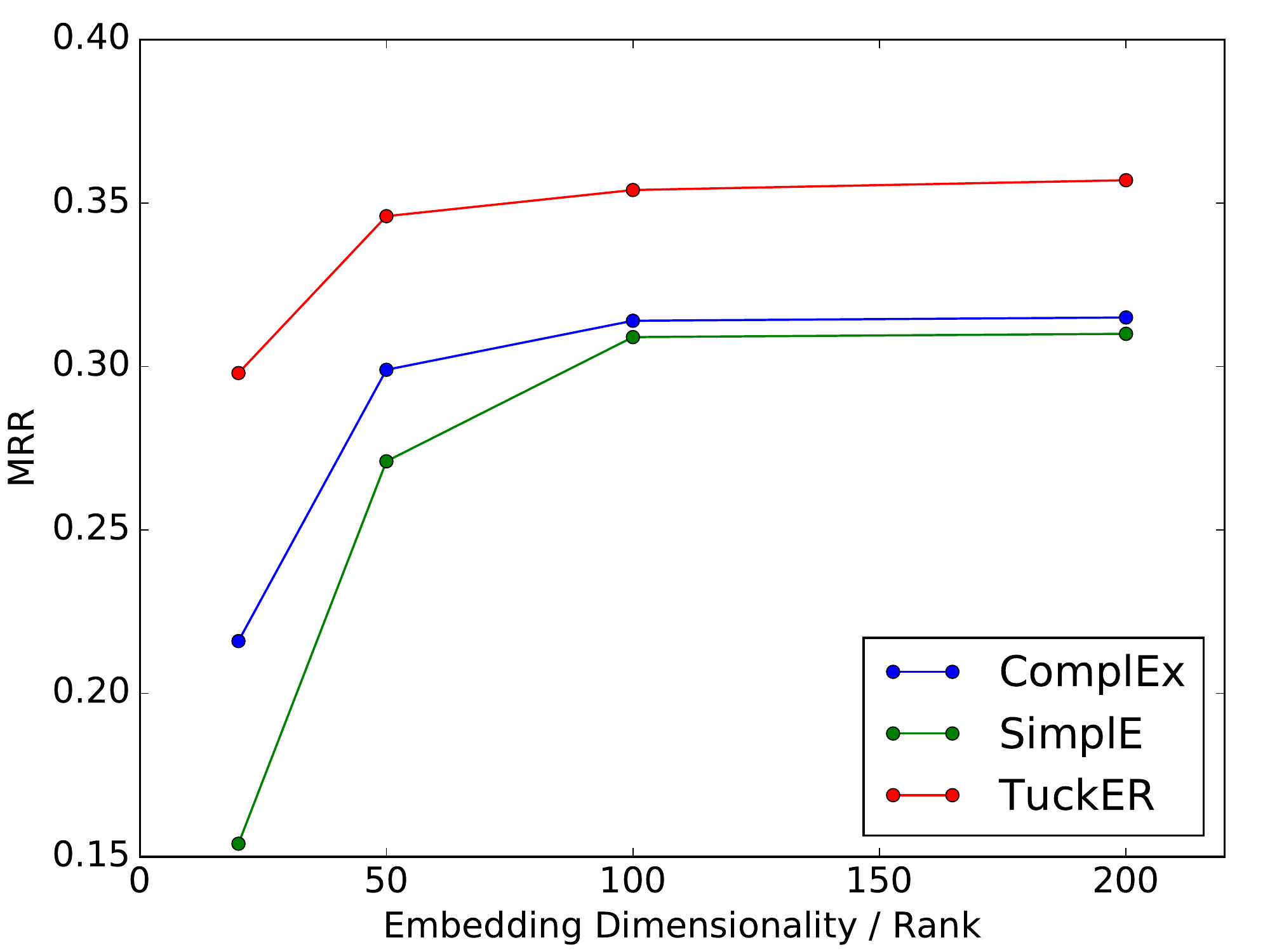}
\caption{MRR for ComplEx, SimplE and TuckER for different embeddings sizes on FB15k-237.}
\label{fig:results}
\end{figure}

The ability of knowledge sharing through the core tensor suggests that TuckER should need a lower number of parameters for obtaining good results than ComplEx or SimplE. To test this, we re-implement ComplEx and SimplE with reciprocal relations, 1-N scoring, batch normalization and dropout for fair comparison, perform random search to choose best hyper-parameters (see Table \ref{table:hyperparams2} in the Appendix \ref{sec:hyperparams} for exact hyper-parameter values used) and train all three models on FB15k-237 with embedding sizes $d_e = d_r \in\{20, 50, 100, 200\}$. Figure \ref{fig:results} shows the obtained MRR on the test set for each model. It is important to note that at embedding dimensionalities 20, 50 and 100, TuckER has fewer parameters than ComplEx and SimplE (e.g. ComplEx and SimplE have $\sim$3 million and TuckER has $\sim$2.5 million parameters for embedding dimensionality 100).

 We can see that the difference between the MRRs of ComplEx, SimplE and TuckER is approximately constant for embedding sizes 100 and 200. However, for lower embedding sizes, the difference between MRRs increases by $0.7\%$ for embedding size 50 and by $4.2\%$ for embedding size 20 for ComplEx and by $3\%$ for embedding size 50 and by $9.9\%$ for embedding size 20 for SimplE. At embedding size 20 ($\sim$300k parameters), the performance of TuckER is almost as good as the performance of ComplEx and SimplE at embedding size 200 ($\sim$6 million parameters), which supports our initial assumption.

\section{Conclusion}
\vspace{-0.1cm}

In this work, we introduce TuckER, a relatively straightforward linear model for link prediction on knowledge graphs, based on the Tucker decomposition of a binary tensor of known facts. TuckER achieves state-of-the-art results on standard link prediction datasets, in part due to its ability to perform multi-task learning across relations. Whilst being fully expressive, TuckER's number of parameters grows linearly with respect to the number of entities or relations in the knowledge graph. We further show that previous linear state-of-the-art models, RESCAL, DistMult, ComplEx and SimplE, can be interpreted as special cases of our model. Future work might include exploring how to incorporate background knowledge on individual relation properties into the existing model.

\subsection*{Acknowledgements}
Ivana Bala\v{z}evi\'c and Carl Allen were supported by the Centre for Doctoral Training in Data Science, funded by EPSRC (grant EP/L016427/1) and the University of Edinburgh.

\bibliography{tucker}
\bibliographystyle{acl_natbib}

\clearpage
\appendix

\section{Hyper-parameters}
\label{sec:hyperparams}

Table \ref{table:hyperparams} shows best performing hyper-parameter values for TuckER across all datasets, where lr denotes learning rate, dr decay rate, ls label smoothing, and d\#$k$, $k \in \{1, 2, 3\}$ dropout values applied on the subject entity embedding, relation matrix and subject entity embedding after it has been transformed by the relation matrix respectively.

\begin{table}[!htbp]
\centering
\resizebox{7.5cm}{!}{
\begin{tabular}{lccccccccc}
  \toprule
  Dataset & lr & dr & $d_e$ & $d_r$ & d\#1  & d\#2 & d\#3 & ls\\
  \midrule
  FB15k & 0.003 & 0.99 & 200 & 200 & 0.2 & 0.2 & 0.3 & 0. \\
  FB15k-237 & 0.0005 & 1.0 & 200 & 200 & 0.3 & 0.4 & 0.5 & 0.1 \\
  WN18 & 0.005 & 0.995 & 200 & 30 & 0.2 & 0.1 & 0.2 & 0.1 \\
  WN18RR & 0.01 & 1.0 & 200 & 30 & 0.2 & 0.2 & 0.3 & 0.1 \\
  \bottomrule
\end{tabular}
}
  \caption{Best performing hyper-parameter values for TuckER across all datasets.}
    \label{table:hyperparams}
\end{table}

Table \ref{table:hyperparams2} shows best performing hyper-parameter values for ComplEx and SimplE on FB15k-237, used to produce the result in Figure \ref{fig:results}.

\begin{table}[!htbp]
\centering
\resizebox{7.5cm}{!}{
\begin{tabular}{lccccccccc}
  \toprule
  Model & lr & dr & $d_e$ & $d_r$ & d\#1  & d\#2 & d\#3 & ls\\
  \midrule
  ComplEx & 0.0001 & 0.99 & 200 & 200 & 0.2 & 0. & 0. & 0.1 \\
  SimplE & 0.0001 & 0.995 & 200 & 200 & 0.2 & 0. & 0. & 0.1 \\
  \bottomrule
\end{tabular}
}
  \caption{Best performing hyper-parameter values for ComplEx and SimplE on FB15k-237.}
    \label{table:hyperparams2}
\end{table}

\end{document}